\documentclass{article} 
\usepackage[final]{colm2025}  

\usepackage[T1]{fontenc}
\usepackage{inconsolata}

\usepackage{microtype}
\usepackage{hyperref}
\usepackage{url}
\usepackage{booktabs}
\usepackage{graphicx}
\usepackage{lineno}
\usepackage{dirtytalk}
\usepackage{amssymb}
\usepackage{amsmath}
\usepackage[capitalise,noabbrev]{cleveref}
\usepackage{listings}
\usepackage{xcolor}
\usepackage{caption}
\usepackage{textcomp}  
\usepackage{subcaption}
\usepackage{graphicx}
\usepackage{adjustbox}
\usepackage{float}

\definecolor{darkblue}{rgb}{0, 0, 0.5}
\definecolor{stringcolor}{RGB}{186,33,33}  
\definecolor{highlightbg}{RGB}{255,255,180}  

\hypersetup{colorlinks=true, citecolor=darkblue, linkcolor=darkblue, urlcolor=darkblue}

\newcommand{\ourmethod}{\textsc{BoundlessBPE}}

\newcommand{\hstring}[1]{\colorbox{highlightbg}{\textcolor{stringcolor}{\textbf{#1}}}}

\DeclareCaptionType{listing}[Listing][List of Listings]

\lstdefinestyle{numberedlist} {
    language=Python,
    basicstyle=\ttfamily\small,
    keepspaces=true,
    columns=flexible,
    numbers=left,
    numberstyle=\tiny\color{gray},
    numbersep=5pt,
    stepnumber=1,
    stringstyle=\color{stringcolor},
    showstringspaces=false,
    escapeinside={@}{@},
}

\lstdefinestyle{smallsize} {
    language=Python,
    basicstyle=\ttfamily\small,
    keepspaces=true,
    columns=flexible,
    stringstyle=\color{stringcolor},
    showstringspaces=false,
    escapeinside={@}{@},
}

\newtoggle{colmfinal}

\title{Boundless Byte Pair Encoding: \\ Breaking the Pre-tokenization Barrier}

\author{Craig W. Schmidt, Varshini Reddy \& Chris Tanner\thanks{Chris Tanner is also affiliated with MIT in Cambridge, MA, USA.} \\
Kensho Technologies\\
Cambridge, MA 02138, USA \\
\texttt{\{craig.schmidt,varshini.bogolu,chris.tanner\}@kensho.com} \\
\And
Yuval Pinter \\
Faculty of Computer and Information Science \\
Ben-Gurion University of the Negev \\
Beer Sheva, Israel \\
\texttt{uvp@cs.bgu.ac.il} \\
}

\begin{document}

\ifcolmsubmission
\linenumbers
\fi

\maketitle

\begin{abstract}
Pre-tokenization, the initial step in many modern tokenization pipelines, segments text into smaller units called \textit{pretokens}, typically splitting on whitespace and punctuation. While this process encourages having full, individual words as tokens, it introduces a fundamental limitation in most tokenization algorithms such as Byte Pair Encoding (BPE). Specifically, pre-tokenization causes the distribution of tokens in a corpus to heavily skew towards common, full-length words. This skewed distribution limits the benefits of expanding to larger vocabularies, since the additional tokens appear with progressively lower counts. To overcome this barrier, we propose \textit{\ourmethod{}}, a modified BPE algorithm that relaxes the pretoken boundary constraint. Our approach selectively merges two complete pretokens into a larger unit we term a \textit{superword}. Superwords are not necessarily semantically cohesive. For example, the pretokens \texttt{\textvisiblespace of} and \texttt{\textvisiblespace the} might be combined to form the superword \texttt{\textvisiblespace of\textvisiblespace the}. This merging strategy results in a substantially more uniform distribution of tokens across a corpus than standard BPE, and compresses text more effectively, with up to a 15\% increase in bytes per token.

\end{abstract}

\section{Introduction}
\label{sec:intro}
Pre-tokenization is a crucial step in preparing text for language models, helping to align token boundaries to meaningful linguistic units. A document is first broken into chunks called \emph{pretokens} using a regular expression,\footnote{See  \cref{sub:pretoken_reg_ex} and \cref{app:pretokenizer_regex} for a discussion of the particular pre-tokenizer regular expressions we used.} which are then tokenized separately.
Each pretoken may be tokenized into two or more subword tokens, or used as a single token that exactly matches the entire pretoken.
Many common words end up as a single token~\citep{reddy2025much}, more than 90\% for the best baselines (\cref{sec:pretokenization}).
Under such circumstances, differences in the tokenizer itself can only manifest in the small percentage of remaining pretokens.
This high overlap in resulting tokens shared across different tokenizers can explain why tokenizers have been found to perform quite similarly on downstream tasks, with no statistically superior approach~\citep{schmidt-etal-2024-tokenization}.
Selecting a larger vocabulary size tends to exacerbate this issue, since the lower-frequency tokens added to the vocabulary later in the training process do not substantially change the resulting token distribution in the training corpus.

\citet{zouhar-etal-2023-tokenization} suggest that tokenizers exhibiting a more uniform distribution of token frequencies across a corpus tend to yield superior performance in language models.
They suggest avoiding tokens which exhibit high frequency while carrying little semantic content, such as individual bytes, as well as very rare tokens that lack sufficient contextual information for effective learning. However, current pre-tokenization methodologies offer limited control over this distribution, as most pretokens are mapped to single tokens, often encompassing common whole words.

With the goal of obtaining a more uniform distribution of tokens, we propose a modification to the standard pre-tokenization approach that still allows us to retain the benefits of contextual cohesion. Specifically, we introduce \emph{superwords}, tokens composed of an $n$-gram of words, to supplement subwords and words in the vocabulary. The notion of superwords extends beyond semantically cohesive units like \texttt{New\textvisiblespace York\textvisiblespace City} often identified through metrics like Pointwise Mutual Information~\citep[PMI;][]{fano1961transmission,church-hanks-1989-word}.
Instead, they serve to distribute the most common words like \texttt{\textvisiblespace the} into a range of common $n$-grams like \texttt{\textvisiblespace the\textvisiblespace car} and \texttt{\textvisiblespace  the\textvisiblespace house}, thereby lowering the frequency of the most common tokens in a corpus.
As we will demonstrate, a substantial number of such common $n$-grams is found by our approach, enabling more effective utilization of larger vocabularies.

In \cref{sec:superword_bpe}, we introduce \ourmethod{}, an extension to the widely adopted Byte Pair Encoding (BPE) algorithm~\citep{sennrich-etal-2016-neural,10.5555/177910.177914}.
The key modification involves incorporating the concept of a \emph{supermerge}.
During the tokenizer training process, we permit the merging of two adjacent pretokens if each is represented by a single token, in which case, the resulting merge is considered a \emph{superword}.
The supermerge operation lends itself to seamless integration into the standard BPE training procedure:
at each step of training, the merge or permissible supermerge that occurs the most frequently is performed.
We also implement a variation of the deletion technique from PickyBPE~\citep{chizhov-etal-2024-bpe} to remove low-frequency intermediate tokens from the vocabulary.
These deletions contribute to a more uniform distribution at the lower end of the frequency spectrum by freeing up space taken by intermediate tokens for use by more common tokens.

The efficacy of our proposed approach is presented in \cref{sec:token_distribution}, where we use an evaluation corpus to compare the token frequency distributions produced by \ourmethod{} and several commonly-used baseline tokenizers.
The results demonstrate that \ourmethod{} yields a more uniform token distribution, consequently achieving a 3-5\% increase in R\'enyi efficiency \citep{zouhar-etal-2023-tokenization} compared to the baselines.
Since our method exhibits improved compression, with a 9-15\% increase in overall bytes per token, it requires fewer tokens for language model inference.

\section{Limitations of pre-tokenizers}
\label{sec:pretokenization}
Pre-tokenization is a processing step shared by most tokenizers and it plays a vital role in how tokenizer vocabularies are formed. Regular expressions commonly used in pre-tokenization permit spaces only as the first byte in a pretoken (for example, \texttt{\textvisiblespace the}), which supports the alignment of the final tokens with word boundaries. While the objective is to generate tokens that correspond closely to meaningful linguistic units, this alignment does not mandate a strict one-to-one mapping between tokens and complete words.
Instead, it prioritizes the prevention of tokens containing parts of two adjacent words, thereby enabling better downstream performance.
For instance, \citet{schmidt-etal-2024-tokenization} demonstrated that entirely omitting pre-tokenization resulted in the poorest downstream performance among 18 evaluated tokenizers.

\begin{figure}[!t]
    \centering
    \includegraphics[width=0.6\linewidth]{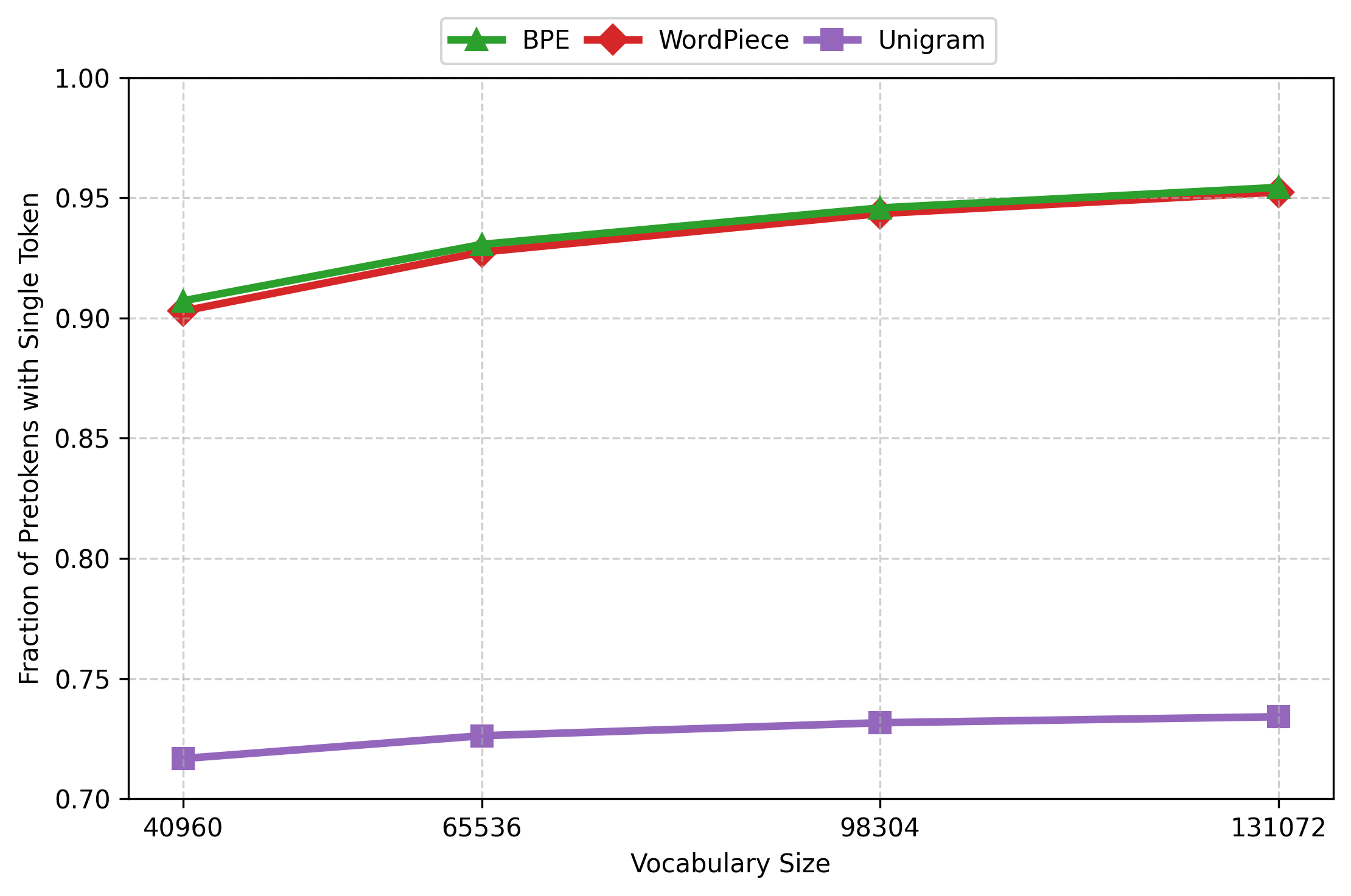}
    \caption{Proportion of pretokens in the evaluation corpus that are represented as full tokens by different tokenization methods (BPE, UnigramLM, WordPiece) across varying vocabulary sizes.}
    \label{fig:pretoken_proportion}
\end{figure}

Due to the Zipfian nature of text, where a few words occur with very high frequency and many words occur rarely, the pretokens of common words are often efficiently represented as a single token that exactly matches the entire pretoken.
We used a 5GB portion of MiniPILE~\citep{kaddour2023minipilechallengedataefficientlanguage} as our out-of-sample evaluation corpus to illustrate this.\footnote{We train tokenizers on the first 170,721 documents of MiniPile, totalling 1GB. The remaining 829,279 documents form the 5GB out-of-sample evaluation corpus used throughout this paper.}
Using the widely adopted BPE, WordPiece ~\citep{wu2016googlesneuralmachinetranslation,wordpiece-org} and UnigramLM~\citep{kudo-2018-subword} tokenizers,\footnote{We use the Hugging Face implementations: \url{https://github.com/huggingface/tokenizers}.} with vocabulary sizes of 40,960, 65,536, 98,304, and 131,072, we calculated the proportion of pretokens that were ultimately represented by a single token.
As shown in \cref{fig:pretoken_proportion}, BPE and WordPiece, both of which are bottom-up merge-based tokenizers, achieve a single-token representation for 90-95\% of pretokens, while UnigramLM, a top-down ablation-based tokenizer, achieves a lower range of 72-73\%.
This highlights how these tokenization methods effectively handle the most frequent words in the vocabulary, and it is a direct consequence of the Zipfian distribution inherent in natural language.

The pre-tokenization step is thus directly determining the vast majority of the tokens, leaving the tokenization training very little ability to modify the token distribution. The phrase \lstinline{'To be or not to be'} composed of very common words will be tokenized by our baseline tokenizers as

\lstinline{['To', ' be', ' or', ' not', ' to', ' be'],}

simply due to pre-tokenization.
To change the tokenization substantially, we need a way to overcome the fact that a pretoken's role during tokenizer training effectively ends once it becomes a single token. For common words, this happens very early in the process.

\section{\ourmethod{}}
\label{sec:superword_bpe}

We introduce \ourmethod{}, a tokenization algorithm that allows adjacent pretokens, each represented as a single token, to be merged into a superword token.
For the Shakespearean example above, our tokenization training process will continue beyond pretoken boundaries, yielding the tokens:\footnote{\ourmethod{} examples throughout this paper are with a vocabulary size of 131,072.}

\lstinline{['To be', ' or not', ' to be'].}

\subsection{Tokenizer training}

Standard BPE training starts with every pretoken in the training data split into individual bytes, which are the initial tokens. For example, \lstinline{'Tip of the hat'} has four pretokens, each containing single-byte initial tokens:

\lstinline{[['T', 'i', 'p'], [' ', 'o', 'f'], [' ', 't', 'h', 'e'], [' ', 'h', 'a', 't']].}

BPE employs pairwise merge rules, such as \lstinline{(' ','t') -> ' t'}, where a token pair is merged to form a single token. At each iteration, the algorithm finds the pair of adjacent tokens with the highest frequency of occurrence across the training data.
This merge rule with the maximum adjacency count is added to a list of merge rules (used during inference), and the newly-combined token is added to the vocabulary.
This pair of tokens is replaced by the combined token throughout the training data, giving:

\lstinline{[['T', 'i', 'p'], [' ', 'o', 'f'], [' t', 'h', 'e'], [' ', 'h', 'a', 't']].}

This process continues until the vocabulary reaches the desired size. During inference, applicable merge rules are applied to a document in the same order they were found in training, until no further merges are possible, giving the final tokenization.

\ourmethod{} is an extension of this standard BPE training process, with two key modifications. First, two adjacent pretokens that are currently tokenized with a single token (identical to the pretoken itself) are allowed to merge into a combined \emph{superword}.
We call these merges \emph{supermerges}.
For example, the adjacent pretokens \lstinline{[' of']} and \lstinline{[' the']} each consisting of a single token:

\lstinline{[['T', 'i', 'p'], [' of'], [' the'], [' h', 'at']],}

can be combined into the superword \lstinline{[' of the']}, which would consist of a single token.
A superword can subsequently be merged with other fully-merged pretokens into longer superwords.\footnote{To avoid combining different pretoken categories, such as digits and words, we restrict the merging process to pretokens matching the regex 
\lstinline{r"^(?=.+\\p\{L\})(?:\\p\{L\}\\p\{M\}*|[ _'\u2019])+\$"}.
This regex matches words with optional initial spaces, contractions, and snake case variables.
See \cref{sub:pretoken_reg_ex} and \cref{app:pretokenizer_regex} for a more detailed explanation of the pre-tokenization regex. Note that numbers are already segmented into groups of 3 in a right-to-left manner, following \citet{singh2024tokenizationcountsimpacttokenization}, so we avoid combining these digit groups into longer superwords to preserve their intended format.  Furthermore, as discussed in \cref{sub:separation_of_unicode}, we opted not to combine punctuation with words.}
Supermerges often involve pairs of common words, such as \lstinline{' of the'}, 
\lstinline{' in the'},
\lstinline{' to the'},
\lstinline{' on the'}, and
\lstinline{' and the'}. One consequence of this behavior is a reduction in the count of the most common tokens such as \lstinline{' the'} and \lstinline{' of'}, as some of their occurrences are allocated to various superwords.

Our second modification to the standard BPE algorithm is the inclusion of a variant of PickyBPE deletions~\citep{chizhov-etal-2024-bpe} to eliminate intermediate tokens that primarily serve as components of more frequent, larger tokens. After a regular merge operation, we use PickyBPE's Intersection over Self (IoS) metric to assess the utility of deletion: we compute the ratio of the frequency with which the merged tokens appear consecutively to the individual frequency of each constituent token.
If the computed IoS value exceeds a threshold, $\tau=0.9$, the constituent token is removed from the vocabulary. This deletion step is effective in eliminating low-count tokens because a token with a high IoS predominantly occurs as part of the newly formed token, thereby freeing up space within the desired vocabulary size for higher-frequency tokens.
For example, after merging \lstinline{' bet'} with \lstinline{'ween'}, the token \lstinline{'ween'} will be deleted, since its IoS is $0.9618 > 0.9$.

For the sake of simplicity, we do not implement deletions following supermerges, as it would require additional bookkeeping of the initial states.\footnote{In our training run with a vocabulary size of $2^{17}$ = 131,072, there would only have been 164 deletions in the 41,038 supermerges, compared to 1,987 deletions for the 89,778 regular merges.
Single bytes are never deleted, as they are necessary to avoid the generation of unknown tokens.}
In the original PickyBPE implementation, a deleted token is split back into the pair that formed it.
We adopt a more aggressive approach where a deleted token is decomposed back into single bytes within a pretoken.
This allows single bytes to potentially recombine into higher-frequency tokens, at the expense of more merge rules.

At any given iteration in the tokenizer training process, we therefore have three potential operations: a regular merge, a supermerge, or a standard deletion.
Both regular merges and supermerges with highest frequency are identified, and the operation with the higher score is performed.
After each regular merge is selected, one or both merged tokens may then be deleted according to their respective IoS values. 

\subsection{Inference procedure}
\label{sub:inference_example}

Once a tokenizer has been trained, it can be used to tokenize a new document, a process often referred to as segmentation or inference. A trained \ourmethod{} tokenizer consists of separate dictionaries for regular merge rules, supermerge rules, and regular deletion rules. Each rule has a unique index representing the order in which it was added during the training process.
Each step of inference involves identifying possible operations that can be performed given the current tokenization of the document. Each pair of adjacent tokens within each pretoken is compared to the merge rules. Each pair of adjacent pretokens containing single tokens is compared to the supermerge rules. Finally, each individual token is compared to the deletion rules.  The operation with the smallest index, meaning it was the first to be added during the training process, is performed everywhere it occurs within the document. \cref{lst:tip_of_the_hat} gives an example of this tokenization process for the phrase \lstinline{'Tip of the hat'}:

\begin{figure}[ht!]
\begin{lstlisting}[style=numberedlist]
[['T', 'i', 'p'], [' ', 'o', 'f'], [' ', 't', 'h', 'e'], [' ', 'h', 'a', 't']]
[['T', 'i', 'p'], [' ', 'o', 'f'], ['@\hstring{ t}@', 'h', 'e'], [' ', 'h', 'a', 't']]
[['T', 'i', 'p'], [' ', 'o', 'f'], [' t', '@\hstring{he}@'], [' ', 'h', 'a', 't']]
[['T', 'i', 'p'], [' ', 'o', 'f'], ['@\hstring{ the}@'], [' ', 'h', 'a', 't']]
[['T', 'i', 'p'], [' ', 'o', 'f'], [' the'], [' ', 'h', '@\hstring{at}@']]
[['T', 'i', 'p'], ['@\hstring{ o}@', 'f'], [' the'], [' ', 'h', 'at']]
[['T', 'i', 'p'], ['@\hstring{ of}@'], [' the'], [' ', 'h', 'at']]
[['T', 'i', 'p'], [' of'], [' the'], ['@\hstring{ h}@', 'at']]
[['T', 'i', 'p'], ['@\hstring{ of the}@'], [' h', 'at']]
[['T', '@\hstring{ip}@'], [' of the'], [' h', 'at']]
[['T', 'ip'], [' of the'], ['@\hstring{ hat}@']]
[['@\hstring{Tip}@'], [' of the'], [' hat']]
\end{lstlisting}
\captionof{listing}{Inference example}
\label{lst:tip_of_the_hat}
\end{figure}

The data is initialized on line 1 with a list of 4 pretokens, each containing a list of individual bytes as the initial tokens. Line 2 shows the results of the first merge of \lstinline{(' ', 't')}, which had the lowest index of any operation. This process of selecting the lowest index merge,  supermerge, or deletion continues until line 7 where \lstinline{[' of']} and \lstinline{[' the']} appear next to each other, and each is a single token. At this point, a supermerge becomes a valid option.  On line 8, a regular merge \lstinline{(' ', 'h')} is performed next due to a lower index, but then the first supermerge creating \lstinline{[' of the']} happens on line 9. The tokenization process concludes on line 12 after the application of a few more merges. 

\subsection{Efficient training implementation}

In the representation shown in \cref{sub:inference_example}, a document is segmented into pretokens, with each pretoken containing one or more tokens. 
While conceptually straightforward and directly applicable to tokenizer training, this representation is computationally inefficient. 
See \cref{app:efficient_implementation} for a description of several techniques used to speed up training time.
We have released an open source version of the \ourmethod{} training and inference code.\footnote{\url{https://github.com/kensho-technologies/boundlessbpe}}

\subsection{Training dynamics}

\begin{figure}[!t]
    \centering
    \begin{minipage}{0.48\textwidth}
        \centering
        \includegraphics[width=\linewidth]{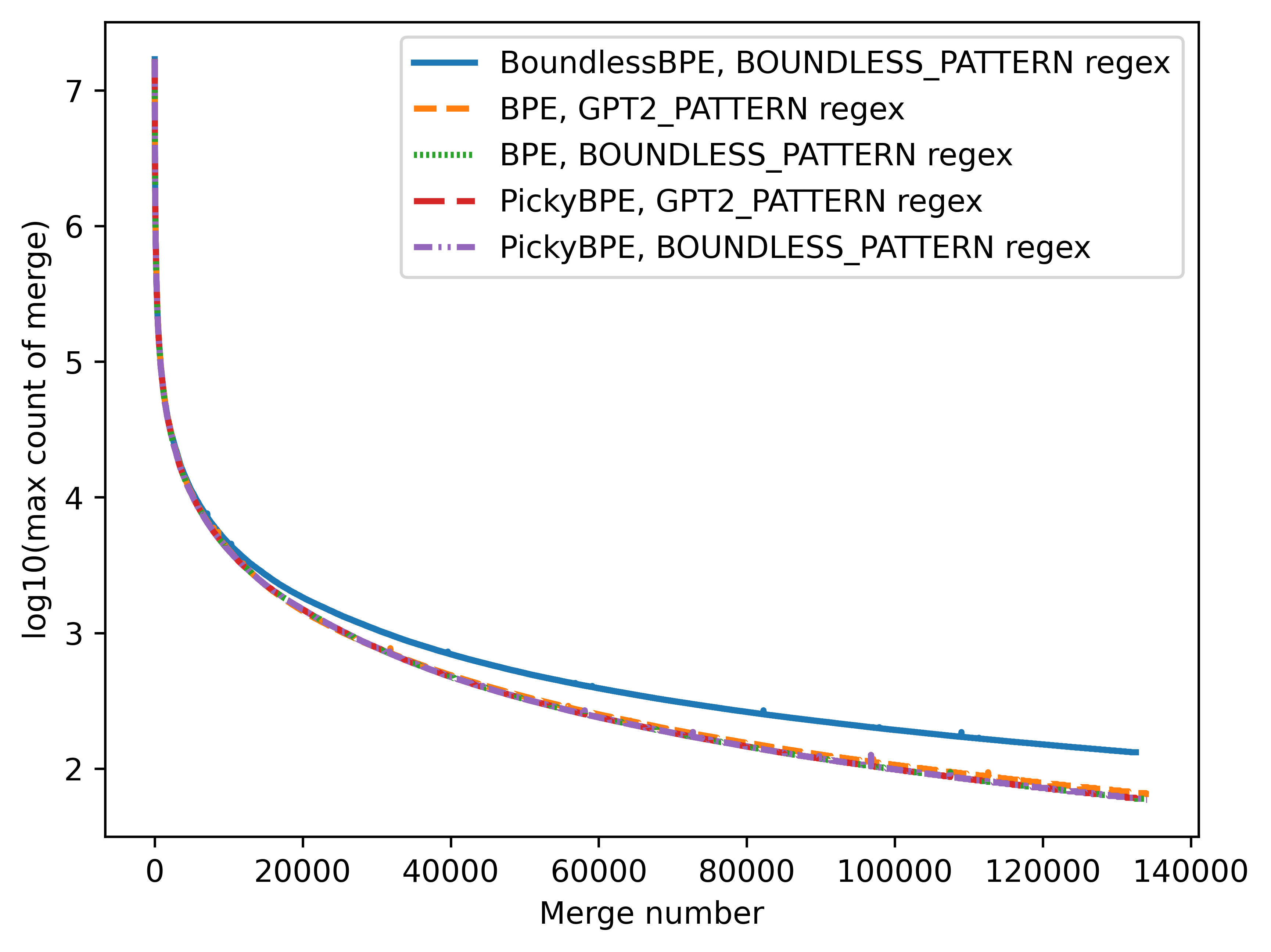}
        \caption{Logarithm of the maximum count of each selected merge, up to a vocabulary size of 131,072.}
        \label{fig:max_cnt}
    \end{minipage}
    \hfill
    \begin{minipage}{0.48\textwidth}
        \centering
        \includegraphics[width=\linewidth]{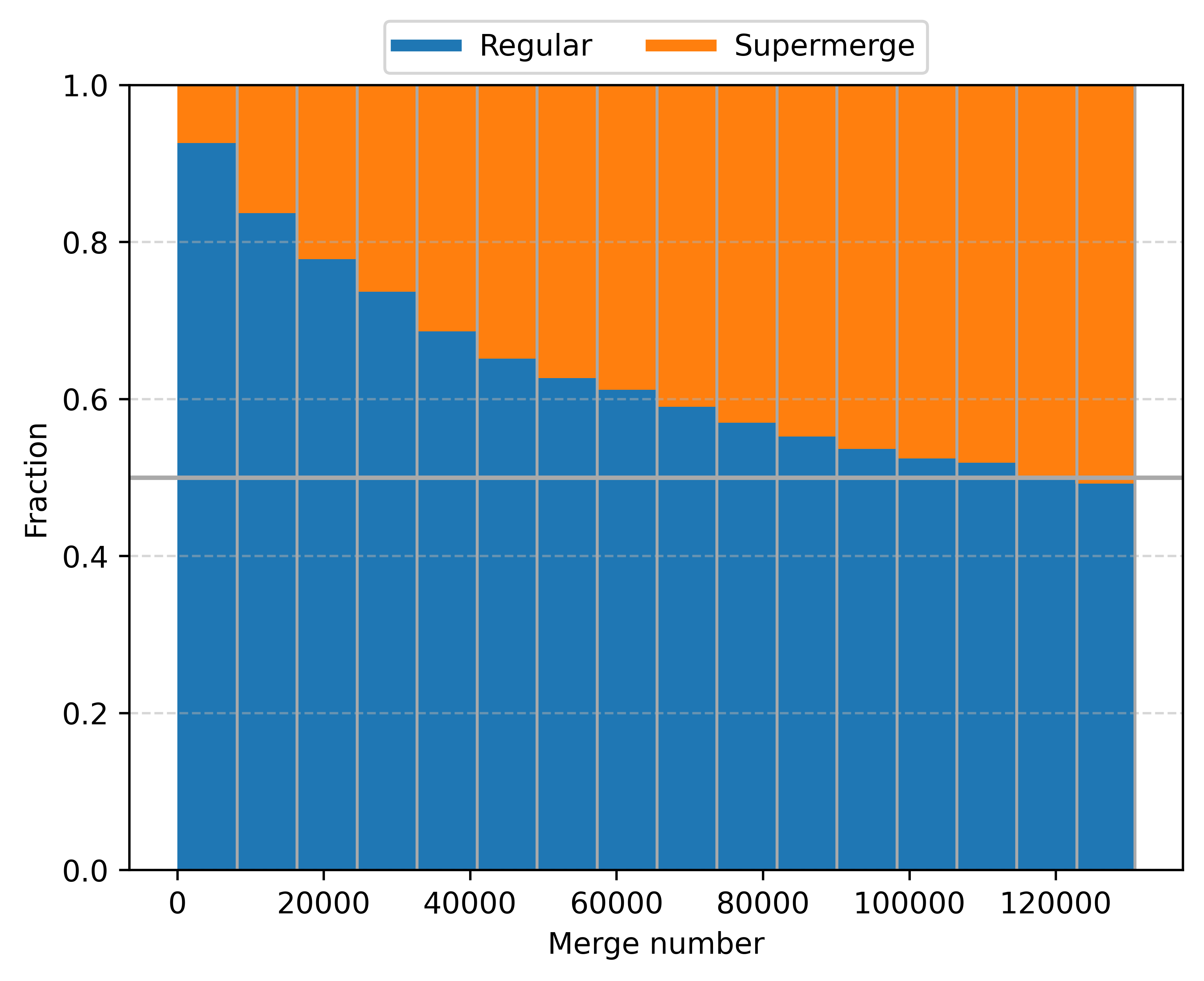}
        \caption{Average fraction of merges (lower) and supermerges (upper) over intervals of 8,192 merges, up to a vocabulary size of 131,072.}
        \label{fig:frac_reg}
    \end{minipage}
\end{figure}

\cref{fig:max_cnt} shows the logarithm of the count of each selected merge or supermerge during the training process for \ourmethod{}, along with the count of each selected merge for standard BPE and PickyBPE, each employing two distinct pre-tokenization regular expressions.
The addition of supermerges as an available choice allows the counts to decrease at a slower rate compared to the baseline methods.
As we will demonstrate, this results in a more uniform token distribution characterized by a tail of tokens with higher frequencies.
The other four curves are essentially indistinguishable, which is a result of so many pretokens ending up as single tokens, as described in \cref{sec:pretokenization}.

\cref{fig:frac_reg} shows the fraction of regular merges and supermerges across each successive group of 8,192 merges.
A supermerge cannot form until both candidate pretokens are a single tokens.
For example, \lstinline{[' of']} and \lstinline{[' the']} need to be represented as single tokens \lstinline{' of'} and \lstinline{' the'}, respectively, before they are eligible for a supermerge.
As a result, early in the process regular merges constitute a majority of operations, and over time the proportion of supermerges grows.
Supermerges were 36.6\% of total merges over the entire range to a vocabulary size of 131,072. In the rightmost interval, supermerges constitute 50.8\% of operations.  Thus, the \ourmethod{} tokenizer uses a substantial number of superwords, breaking through the pre-tokenization barrier. 

\subsection{Superwords enable improved pre-tokenization}
\label{sub:pretoken_reg_ex}

The \texttt{BOUNDLESS\_PATTERN} regex in \cref{lst:ourregex} contains a number of improvements over existing pre-tokenization regex patterns.\footnote{We discuss existing regular expressions and compare them to ours in \cref{app:pretokenizer_regex}.}
One notable improvement is the better handling of names in code, which is directly enabled by the supermerges in \ourmethod{}.
The variable name \lstinline{'XMLHttpRequest'} is composed of three sub-components: \lstinline{'XML'}, \lstinline{'Http'}, and \lstinline{'Request'}, which can be identified by capitalization conventions.
The B-2 to B-6 branches of \cref{lst:ourregex} work together to break variable and function names into smaller pretokens based on capitalization.
These sub-components can then be recombined via supermerges as their co-occurrence counts warrant.
The example \lstinline{'XMLHttpRequest snake_case camelCase CONSTANT'} is pre-tokenized as:

\lstinline{['XML', 'Http', 'Request', ' snake', '_case', ' camel', 'Case', ' CONSTANT'].}

Using the commonly used \texttt{GPT2\_PATTERN}\,\footnotemark[\value{footnote}] pre-tokenization segments on the full names, unable to utilize the prior knowledge encoded in the sub-components:

\lstinline{['XMLHttpRequest', ' snake', '_', 'case', ' camelCase', ' CONSTANT'].}

Conversely, using \texttt{BOUNDLESS\_PATTERN} with standard BPE would always result in the final tokens being the individual sub-components.  With the supermerges \ourmethod{ provides}, it is possible to both align with sub-components and to recombine some of the full names.

\begin{figure}[ht!]
\begin{lstlisting}[style=smallsize]
BOUNDLESS_PATTERN = "|".join([
  r" ?(?:\p{L}\p{M}*)+['\u2019](?:\p{L}\p{M}*)+", # B-1, contraction
  r"_(?:\p{Ll}\p{M}*)+",                          # B-2, snake_case
  r" ?(?:\p{Lu}\p{M}*)+(?=(?:\p{Lu}\p{M}*)(?:\p{Ll}\p{M}*))", # B-3, words
  r" ?(?:\p{Lu}\p{M}*)?(?:\p{Ll}\p{M}*)+",        # B-4, words
  r" ?(?:\p{Lu}\p{M}*)+",                         # B-5, words
  r" ?(?:[\p{Lt}\p{Lm}\p{Lo}]\p{M}*)+",           # B-6, words
  r"(?:\p{N}\p{M}*){1,3}(?=(?:(?:\p{N}\p{M}*){3})*(?:(?:\P{N}\p{M}*)|$))",# B-7
  r" ?(?:[\p{P}\p{S}]\p{M}*)+",                   # B-8, punct and symbols
  r"[^\S\r\n]*[\n\r]+|[^\S\r\n]+",                # B-9, whitespace
  r"(?:[\p{Z}\p{C}]\p{M}*)+",                     # B-10, sep or control
  r"\p{M}+"                                       # B-11, leftover marks
])
\end{lstlisting}
\captionof{listing}{Pre-tokenization regular expressions (regex)}
\label{lst:ourregex}
\end{figure}

\section{Token distribution}
\label{sec:token_distribution}
We have seen that pre-tokenization influences more than 90\% of the tokens in the vocabulary with BPE, which makes it difficult for algorithms like BPE to alter the distribution of token occurrences. However, supermerges offer a means to overcome this barrier.

\begin{figure}[!ht]
    \centering
    \includegraphics[width=\linewidth]{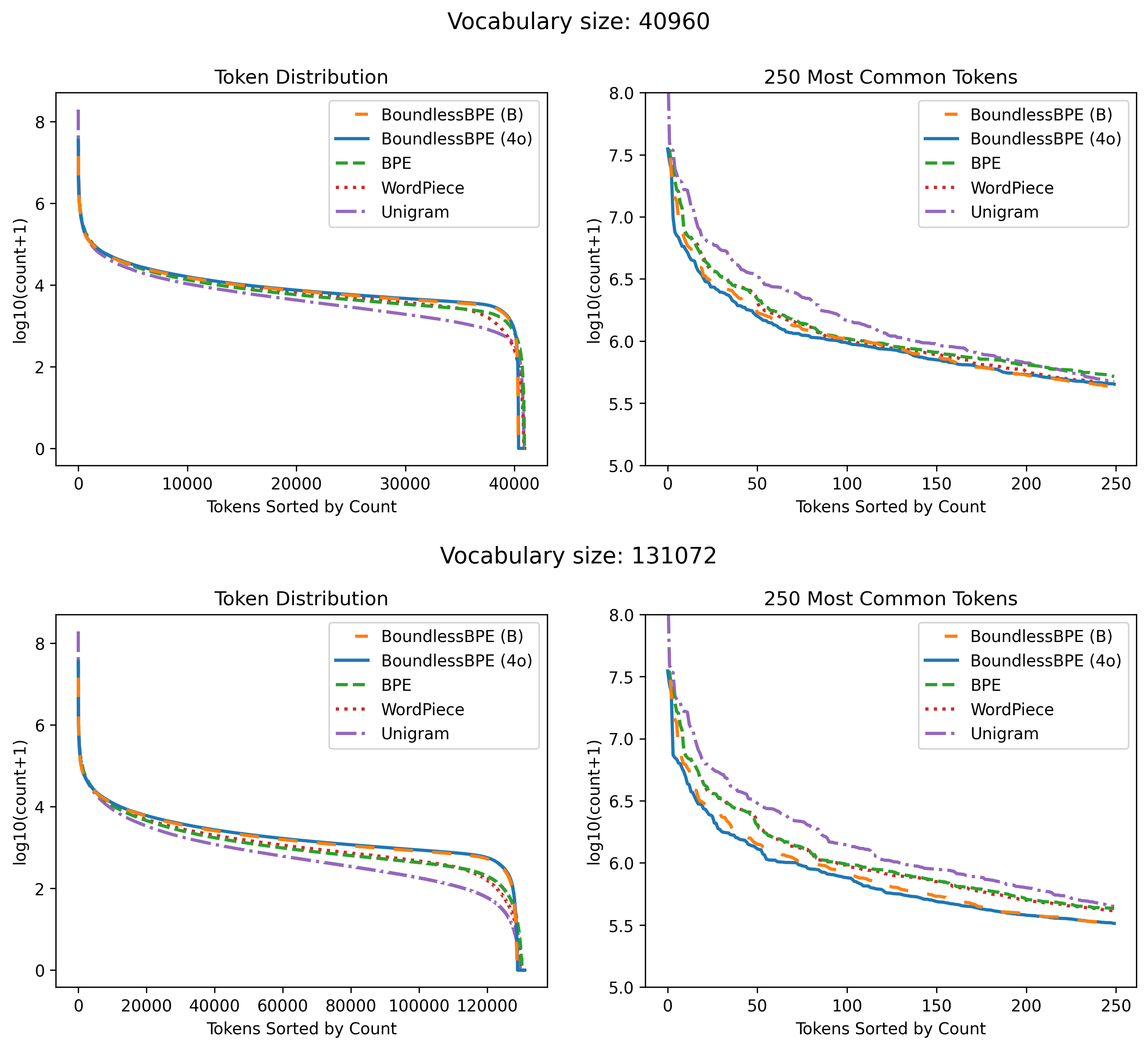}
    \caption{The left column is the $\log_{10}(\text{count}+1)$ for each token, sorted from most to least frequent on the $x$-axis on our evaluation corpus. The +1 is to allow plotting of 0 counts. The right column shows a zoomed-in view of the 250 most common tokens.}
    \label{fig:token_distribution}
\end{figure}

The left column of \cref{fig:token_distribution} shows the log counts for each token in our evaluation corpus, sorted in descending order of frequency along the $x$-axis.
We observe that the tail of \ourmethod{}'s token frequency chart is substantially higher than that of the three baseline tokenizers\footnote{ BoundlessBPE (B) uses the \texttt{BOUNDLESS\_PATTERN} regex, while BoundlessBPE (4o) uses \texttt{GPT4O\_PATTERN} regex, as described in \cref{app:pretokenizer_regex}. The baselines use the \texttt{GPT4O\_PATTERN} regex.} over two vocabulary sizes,\footnote{See \cref{fig:extra_vocab_sizes} in \cref{app:extra_vocab_sizes} for two additional vocabulary sizes.}.
This is due to a combination of higher counts for final merges mentioned above and the deletion of intermediate tokens, which makes room for additional useful tokens.
The right column of \cref{fig:token_distribution} presents a zoomed-in view of this same measure for the 250 tokens with the largest counts in each vocabulary, showing that supermerges have succeeded in reducing the counts of overly-general tokens.

\begin{figure}[!ht]
    \centering
    \includegraphics[width=0.6\linewidth]{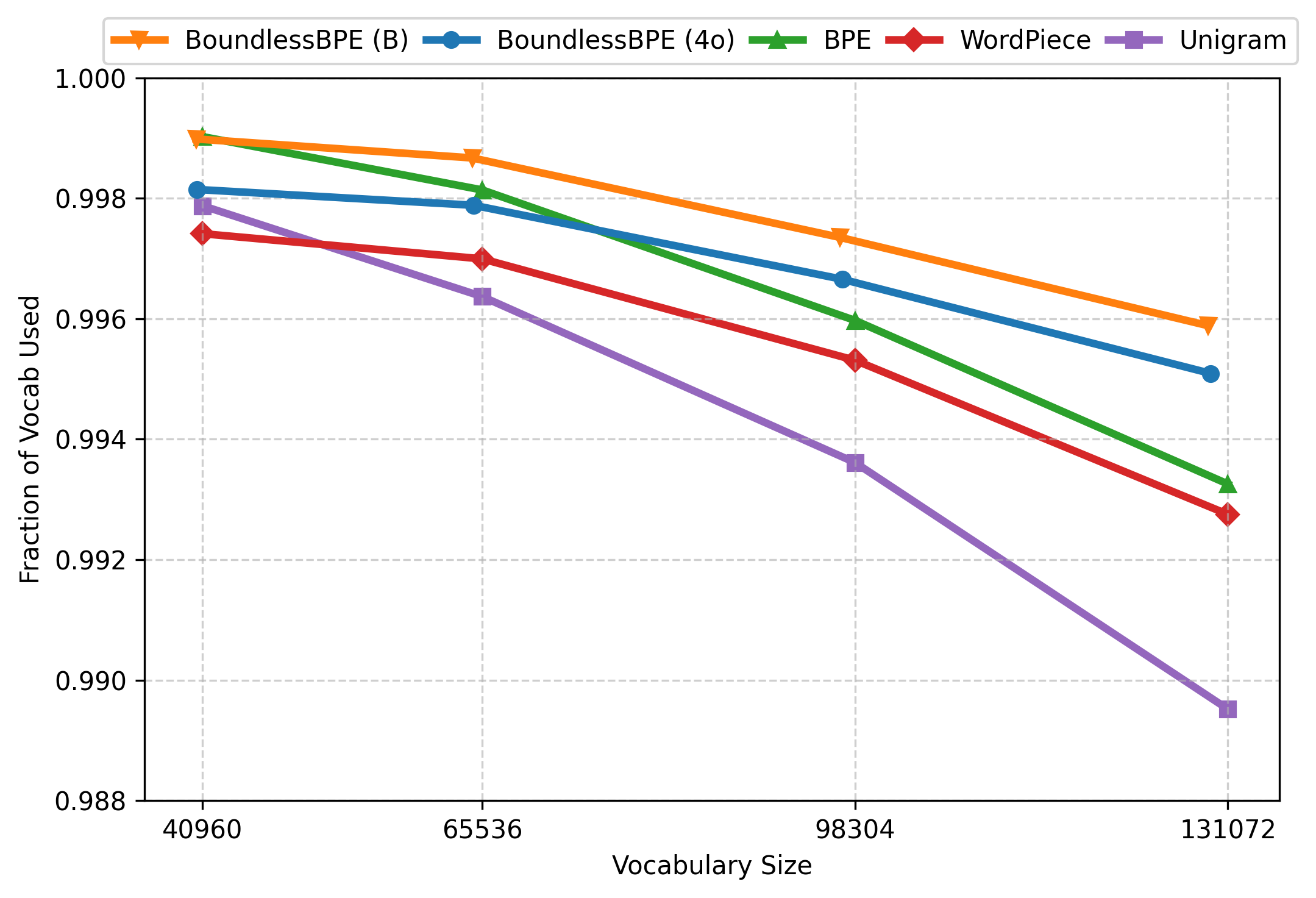}
    \caption{Fraction of vocabulary used at least once in an evaluation corpus, across different tokenization methods and vocabulary sizes. A higher fraction suggests the vocabulary has more useful tokens for representing unseen data.}
    \label{fig:vocab_percent}
\end{figure}

\begin{figure}[htbp]
    \centering
    \begin{minipage}{0.48\textwidth}
        \centering
        \includegraphics[width=\linewidth]{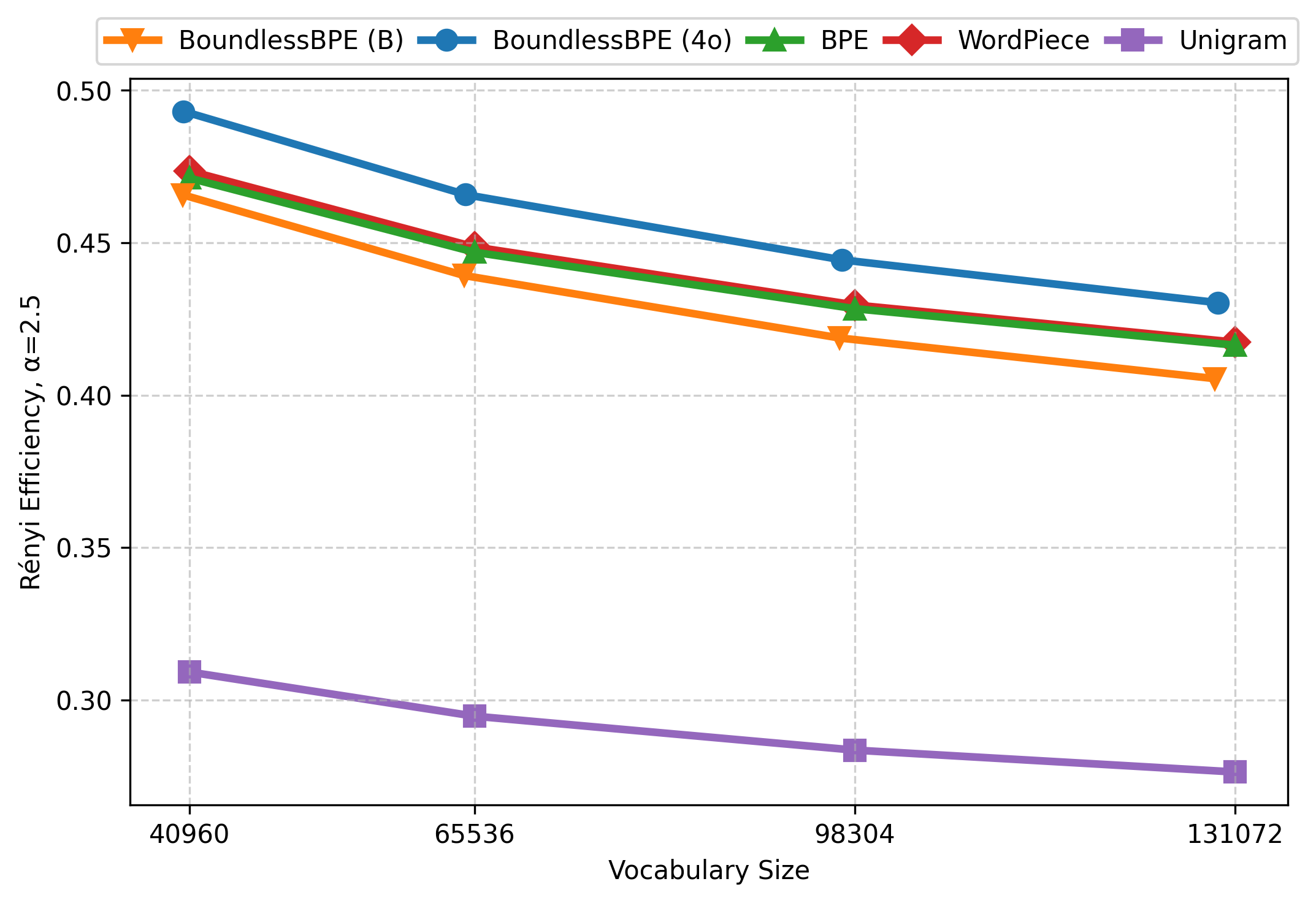}
        \caption{R\'enyi efficiency, calculated over evaluation corpus, with $\alpha$ = 2.5. Tokenizers with higher efficiency are generally desirable.}
        \label{fig:renyi_entropy}
    \end{minipage}
    \hfill
    \begin{minipage}{0.48\textwidth}
        \centering
        \includegraphics[width=\linewidth]{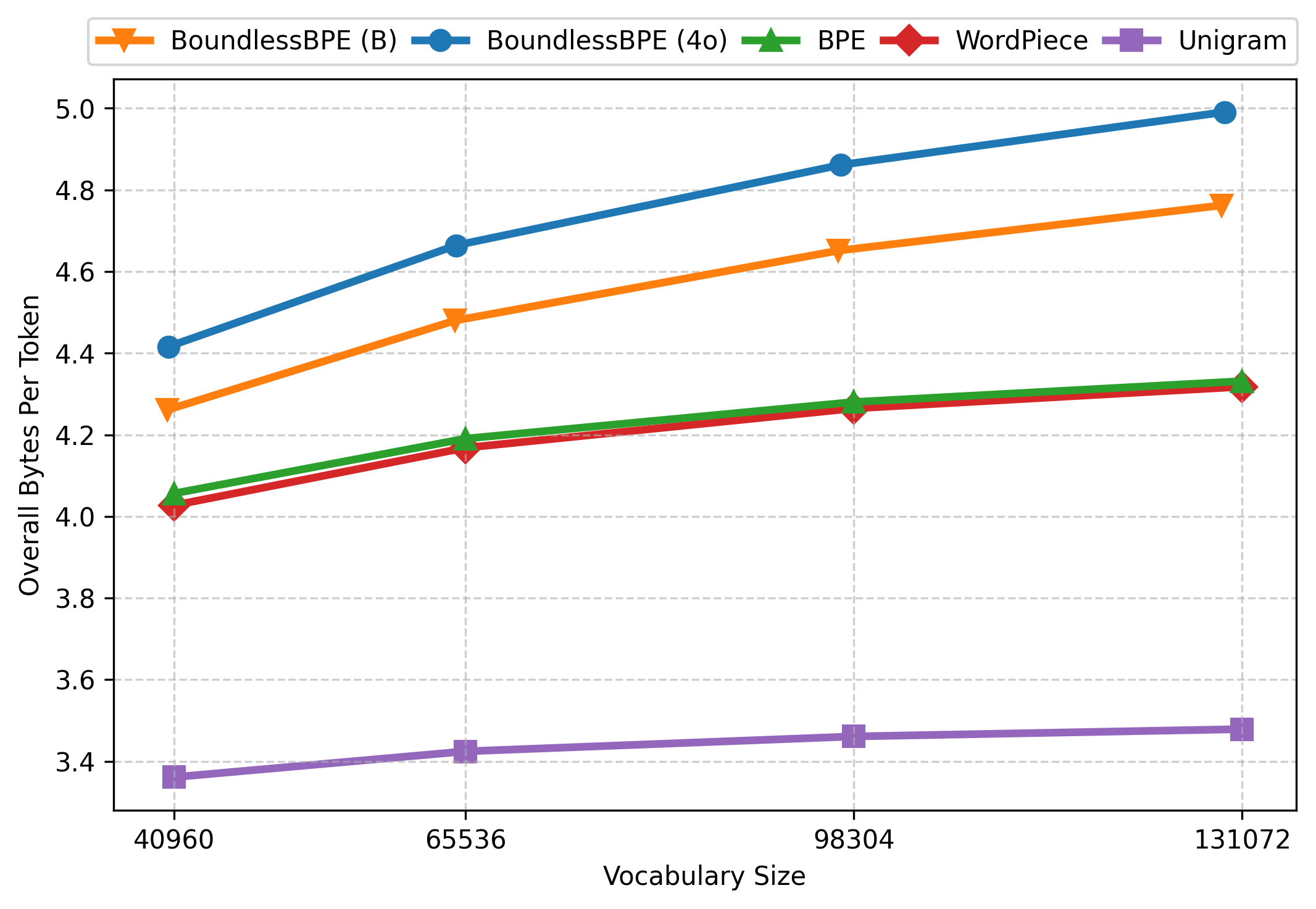}
        \caption{Bytes per token, calculated over evaluation corpus. A higher value implies better compression, which can lead to efficient training and faster inference.}
        \label{fig:compression_ratio_ood}
    \end{minipage}
\end{figure}

\cref{fig:vocab_percent} shows the fraction of the vocabulary that is used at least once on the same evaluation corpus. The \ourmethod{} (B) run did best across vocabulary sizes, with \ourmethod{} (4o) also doing well.\footnote{Note that \cref{fig:vocab_percent} to \cref{fig:compression_ratio_ood} use the effective vocabulary size for \ourmethod{}, subtracting the number of deletions, to make a more accurate comparison.}

Finally, we quantify the token distributions over the corpus using two metrics: (1) the R\'enyi efficiency metric~\citep{zouhar-etal-2023-tokenization} which indicates how uniform a token distribution is; and (2) compression rate of the evaluation corpus.
\cref{fig:renyi_entropy} shows that \ourmethod{} (4o) achieves a R\'enyi efficiency at least 3\% above that of the baselines.\footnote{We use $\alpha = 2.5$ as recommended by \citet{zouhar-etal-2023-tokenization}.} \ourmethod{} (B) had a lower R\'enyi efficiency, showing the sensitivity of this metric to the regex and value of $\alpha$.\footnote{See \cref{app:renyi_analysis} for results at other $\alpha$ values.} There is later work pointing out limitations of R\'enyi efficiency in predicting  downstream performance \citep{cognetta-etal-2024-two}. However, our goal was to design a system with a more uniform distribution of tokens, and it does provide a quantitative measure of the uniformity.

\cref{fig:compression_ratio_ood} shows that \ourmethod{} (4o) provides a 9-15\% increase in overall bytes per token for the 5GB evaluation corpus compared to BPE and WordPiece.
The compression rate continues to increase at larger vocabularies, indicating it is able to effectively use the additional vocabulary space.
While the effect of compression on downstream performance is unclear~\citep{galle-2019-investigating,goldman-etal-2024-unpacking,ali-etal-2024-tokenizer,schmidt-etal-2024-tokenization}, having more bytes per token can speed up language model inference, as fewer tokens are needed to process or generate a given text.

\section{Ablation study}
\label{sec:ablations}

\ourmethod{} combined supermerges with a new pre-tokenization regular expression, \texttt{BOUNDLESS\_PATTERN}. 
We also used a different form of PickyBPE deletions, where deleted tokens were split into individual bytes before being allowed to re-merge. Our motivation was that the more aggressive decomposition could allow the deleted token to form more compressed tokens.
In contrast, \citet{chizhov-etal-2024-bpe} break a deleted token into the tokens that created it in a pairwise merge. 

\begin{table}[htbp]
\centering
\begin{tabular}{cllcc}
\toprule
\textbf{Vocab Size} & \textbf{Regex Pattern} & \textbf{PickyBPE} & \textbf{Bytes Per Token} & \textbf{R\'enyi Efficiency} \\
\midrule
40,960 & Boundless & none & 4.268 & 0.4650 \\
40,960 & Boundless & original & 4.267 & 0.4650 \\
40,960 & Boundless & ours & 4.261 & 0.4651 \\
\midrule
40,960 & GPT4o & none & 4.424 & 0.4925 \\
40,960 & GPT4o & original & 4.422 & 0.4925 \\
40,960 & GPT4o & ours & 4.416 & 0.4926 \\
\midrule
131,072 & Boundless & none & 4.771 & 0.4048 \\
131,072 & Boundless & original & 4.769 & 0.4049 \\
131,072 & Boundless & ours & 4.762 & 0.405 \\
\midrule
131,072 & GPT4o & none & 5.001 & 0.4298 \\
131,072 & GPT4o & original & 4.999 & 0.4299 \\
131,072 & GPT4o & ours & 4.991 & 0.4300 \\
\midrule

\bottomrule
\end{tabular}
\caption{Effect of regular expression pattern and PickyBPE style on intrinsic measures}
\label{tab:ablation}
\end{table}

\cref{tab:ablation} gives an ablation study that disentangles these changes.\footnote{See \cref{app:ablation_extra_vocab_sizes} for the same results in two additional vocabulary sizes.}
We consider both forms of PickyBPE deletions, as well as no deletions, and compare our \texttt{BOUNDLESS\_PATTERN} to a modern baseline of the GPT4o regular expression.\footnote{See \cref{app:pretokenizer_regex}}  
We see that the differences in the types of deletions had only a very small effect on either of our intrinsic measures. 
Still, deletions may prove important for downstream performance, since low-frequency tokens are harder to learn \citep{su-etal-2024-mile,yu-etal-2022-rare}.
Substituting the regex for the GPT4o version increased bytes per token by 0.15-0.23, and increased R\'enyi efficiency by around 0.025. The \texttt{GPT4O\_PATTERN} discussed in \cref{app:pretokenizer_regex} contains \texttt{O-1} and \texttt{O-2}, which allow a single non-alphanumeric character to combine with word tokens, compressing more than \texttt{BOUNDLESS\_PATTERN}, which was designed to keep character classes more separate.

\section{Related work}
\label{sec:related_work}
\paragraph{\textbf{Impact of pre-tokenization}}
\citet{velayuthan-sarveswaran-2025-egalitarian} emphasize the importance of pre-tokenization relative to the tokenizer in achieving egalitarian tokenization across languages.
They observe that pre-tokenization limits achievable compression, since each pretoken must contain at least one token.
Thus, the number of of pretokens is a lower bound on the number of tokens.
\citet{pmlr-v235-dagan24a} also showed the substantial impact of pre-tokenization regex choices on tokenizer compression and downstream performance.
Furthermore, \citet{wegmann2025tokenizationsensitivelanguagevariation} demonstrate that pre-tokenization has a stronger impact on downstream task performance than that of  vocabulary size and training corpus variations.

\paragraph{\textbf{Multi-word tokens}}
Prior work has explored incorporating larger linguistic units into vocabularies.
\citet{salehi-etal-2015-word,liu2025superbpespacetravellanguage, huang2025overtokenizedtransformervocabularygenerally} highlight the benefits of multi-word tokens for compression, training cost, and model performance. Similarly, \citet{otani-etal-2020-pre} show representation improvements using multi-word expressions (MWE's) in multilingual settings.
\citet{kumar-thawani-2022-bpe} found that adding high-PMI MWE's improved performance of machine translation better than high-frequency subword or whole-word tokens.
\citet{gee-etal-2023-multi} introduced a multi-word tokenizer that augments a standard BPE vocabulary with MWE's by representing frequent $n$-grams as a single token.

In concurrent work, \citet{liu2025superbpespacetravellanguage} proposed SuperBPE, an enhancement to BPE that employs a two-pass tokenization strategy to obtain multi-word tokens, which they also term \emph{superwords}.
Their method involves an initial BPE training phase with pre-tokenization, conducted up to a vocabulary size $t < T$, where $T$ represents a hyperparameter they call the \emph{transition point}. 
This phase is followed by a second BPE training pass that resumes from the first but omits pre-tokenization, thereby enabling the formation of superwords to populate the remainder of the vocabulary.
In contrast to SuperBPE, \ourmethod{} operates in a single pass rather than in separate stages, allowing both standard merges and supermerges to occur at the same points based on their respective frequencies.
Our approach thus obviates the need for a transition point, avoiding a hyperparameter search and speeding up tokenizer training.
Additionally, our method offers control over which pretokens can be merged together, preventing different types of pretokens, such as words and punctuation, from merging.

See \cref{app:additional_related_work} for additional related work.

\section{Conclusion}
\label{sec:conclusion}
While natural language processing has achieved significant advancements in performance over the past decade, certain design choices remain static, such as tokenization algorithms and pre-tokenization regular expressions. Pre-tokenization exerts considerable influence over a corpus token distribution, with standard pre-tokenization methods fixing over 90\% of words to be represented as single tokens. To address the current limitations in achieving a uniform token distribution over a corpus, we introduce two key contributions. First, we present \ourmethod{}, a modified BPE training process that enables the combination of adjacent full pre-tokens into superwords. The incorporation of superwords yields enhanced compression, quantified by an increase in bytes per token, and a more uniform distribution of token frequencies across a corpus. Second, we propose \texttt{BOUNDLESS\_PATTERN}, a novel pre-tokenization regular expression designed to work with superwords, resulting in improved tokenization, particularly for code and named entities, when compared to standard regular expressions such as the one used by GPT-2. Based on prior work \citep{zouhar-etal-2023-tokenization,liu2025superbpespacetravellanguage}
we hypothesize that the intrinsic performance improvements shown by \ourmethod{} will have a positive impact on the downstream performance of language models.

\section*{Acknowledgments}
Many thanks to Ilya Yudkovich and Mike Arov at Kensho Technologies for their technical assistance. 
Thanks also to Seth Ebner, Charles Lovering, and Michael Krumdick at Kensho Technologies for many helpful comments and discussions.
This research was supported in part by the Israel Science Foundation (grant No. 1166/23).

\bibliography{anthology,colm2025}
\bibliographystyle{colm2025}

\appendix
\section{Pre-tokenizer regular expression}
\label{app:pretokenizer_regex}

Pre-tokenization applies a regular expression (regex) to each document to form pretokens, which are then each tokenized separately. The regex in this section make extensive use of Unicode categories.  These are not supported by the \lstinline{re} library used by default in python, instead requiring the more powerful \lstinline{regex} library.\footnote{\url{https://pypi.org/project/regex/}}  In the notation of \lstinline{regex}, Unicode can be divided into the categories given in \cref{tab:unicode-categories}.

\begin{table}[htbp]
\centering
\begin{tabular}{ll}
\hline
\textbf{Category} & \textbf{Description} \\
\hline
\lstinline|\p{L}| & Letters \\
\lstinline|\p{N}| & Numbers \\
\lstinline|\p{Z}| & Whitespace and other separators \\
\lstinline|\p{S}| & Symbols \\
\lstinline|\p{P}| & Punctuation \\
\lstinline|\p{C}| & Control characters \\
\lstinline|\p{M}| & Combining marks (diacritical marks, etc.) \\
\hline
\end{tabular}
\caption{Unicode Character Categories}
\label{tab:unicode-categories}
\end{table}

Any regular expression can be used for pre-tokenization, provided that it matches all of the text in a given Unicode string. Thus, at least one branch of the regex must match each of these Unicode categories. 

Previous regex were used without much emperical justification.  \citet{pmlr-v235-dagan24a} and \citet{wegmann2025tokenizationsensitivelanguagevariation} are among the first to more systematically investigate the effect of the regex on downstream performance.

\cref{lst:regex} gives the regex for GPT-2 , GPT-4,\footnote{Taken from \url{https://github.com/karpathy/minbpe}}, GPT-4o,\footnote{\url{https://github.com/openai/tiktoken/blob/4560a889/tiktoken\_ext/openai\_public.py\#L101-L114}} and the Punct regex \citep{pmlr-v235-dagan24a}. The proposed regex for \ourmethod{} is given in \cref{lst:boundlessregex}. Each of the regular expression branches shown in the labeled lists are combined together with the \lstinline{|} operator into a single regex. In many of the regex, an initial \lstinline|r" ?"| indicates an optional initial space, while a final \lstinline{r"[\r\n]*"} indicates zero or more line endings.

\begin{figure}[ht!]
\begin{lstlisting}
import regex as re

GPT2_PATTERN = "|".join([
    r"'(?:[sdmt]|ll|ve|re)",        # T-1, English contractions
    r" ?\p{L}+",                    # T-2, words
    r" ?\p{N}+",                    # T-3, digits
    r" ?[^\s\p{L}\p{N}]+",          # T-4, not letters, digits, or whitespace
    r"\s+(?!\S)",                   # T-5, all-but-last whitespace
    r"\s+"                          # T-6, whitespace
])

GPT4_PATTERN = "|".join([
    r"'(?i:[sdmt]|ll|ve|re)",       # F-1, English contractions
    r"[^\r\n\p{L}\p{N}]?+\p{L}+",   # F-2, words, w/ opt non-alphanumeric
    r"\p{N}{1,3}",                  # F-3, digits
    r" ?[^\s\p{L}\p{N}]++[\r\n]*",  # F-4, not letters, digits, or whitespace
    r"\s*[\r\n]",                   # F-5, whitespace with line-ending
    r"\s+(?!\S)",                   # F-6, all-but-last whitespace
    r"\s+"                          # F-7, all whitespace
])

GPT4O_PATTERN = "|".join([
    r"[^\r\n\p{L}\p{N}]?[\p{Lu}\p{Lt}\p{Lm}\p{Lo}\p{M}]*\
[\p{Ll}\p{Lm}\p{Lo}\p{M}]+\
(?i:'s|'t|'re|'ve|'m|'ll|'d)?",     # O-1 word with some lowercase
    r"[^\r\n\p{L}\p{N}]?[\p{Lu}\p{Lt}\p{Lm}\p{Lo}\p{M}]+\
[\p{Ll}\p{Lm}\p{Lo}\p{M}]*\
(?i:'s|'t|'re|'ve|'m|'ll|'d)?",     # O-2 word with some uppercase
    r"\p{N}{1,3}""",                # O-3, digits
    r" ?[^\s\p{L}\p{N}]+[\r\n/]*",  # O-4, not letters, digits, or whitespace
    r"\s*[\r\n]+",                  # O-5, whitespace with line-ending
    r"\s+(?!\S)",                   # O-6, all-but-last whitespace
    r"\s+",                         # O-7, all whitespace
])

PUNCT_PATTERN = "|".join([
    r" ?\p{L}+",                    # P-1, words
    r"\p{N}{1,3}",                  # P-2, digits
    r" ?[^\s\p{L}\p{N}]+[\r\n]*",   # P-3, not letters, digits, or whitespace
    r"\s*[\r\n]+",                  # P-4, whitespace with line-ending
    r"\s+(?!\S)",                   # P-5, all-but-last whitespace
    r"\s+"                          # P-6, whitespace
])

\end{lstlisting}
\captionof{listing}{Comparison of existing pre-tokenization regular expressions (regex)}
\label{lst:regex}
\end{figure}

\begin{figure}[ht!]
\begin{lstlisting}
BOUNDLESS_PATTERN = "|".join([
    # B-1, contraction, allow curly apostrophe
    r" ?(?:\p{L}\p{M}*)+['\u2019](?:\p{L}\p{M}*)+", 
    # B-2, snake_case, with underscore at front
    r"_(?:\p{Ll}\p{M}*)+",                          
    # B-3, Uppercase, followed by uppercase and lowercase letter
    r" ?(?:\p{Lu}\p{M}*)+(?=(?:\p{Lu}\p{M}*)(?:\p{Ll}\p{M}*))", 
    # B-4, optional uppercase, and one or more lowercase
    r" ?(?:\p{Lu}\p{M}*)?(?:\p{Ll}\p{M}*)+",        
    # B-5, all uppercase acronym CONSTANT
    r" ?(?:\p{Lu}\p{M}*)+",                         
    # B-6, titlecase, modifier, or uncased letters
    r" ?(?:[\p{Lt}\p{Lm}\p{Lo}]\p{M}*)+",           
    # B-7, numbers
    r"(?:\p{N}\p{M}*){1,3}(?=(?:(?:\p{N}\p{M}*){3})*(?:(?:\P{N}\p{M}*)|$))",  
    # B-8, optional space, punctuation and symbols
    r" ?(?:[\p{P}\p{S}]\p{M}*)+",                     
    # B-9, whitespace
    r"[^\S\r\n]*[\n\r]+|[^\S\r\n]+",                
    # B-10, separator or control
    r"(?:[\p{Z}\p{C}]\p{M}*)+",                     
    # B-11, leftover marks, just for bad utf-8
    r"\p{M}+"                              
])    
\end{lstlisting}
\captionof{listing}{BoundlessBPE regular expression (regex)}
\label{lst:boundlessregex}
\end{figure}

\subsection{Separation of Unicode Classes}
\label{sub:separation_of_unicode}

One open question with the pre-tokenization regex is how much the Unicode character classes should be kept separate. The \texttt{GPT2\_PATTERN} pattern kept them very separate.  The \texttt{GPT4\_PATTERN} pattern moved away from that with pattern in line \texttt{F-2} that allowed any single character besides a letter, digit, or line ending to come before a word.  It also combined line endings with other characters in \texttt{F-4}.  The \texttt{GPT4O\_PATTERN} is similar, further breaking words based on case and keeping contractions together. To study if separate Unicode character classes are good idea, the Punct pattern \citep{pmlr-v235-dagan24a} returned to only allowing a space before a word with \texttt{P-1}.  They found an improvement for \texttt{PUNCT\_PATTERN} over \texttt{GPT4\_PATTERN} at a vocabulary size of 32k, but no significant difference at a vocabulary size of 80k.  \citet{wegmann2025tokenizationsensitivelanguagevariation} also had mixed results on this question.  They found the \texttt{GPT2\_PATTERN} with the highest separation was best for tasks requiring robustness to language variation, while more mixing was beneficial to tasks requiring sensitivity to language variation.

In the face of mixed evidence, we keep our classes well separated, in the hope that this will give more of the word based tokens that can be combined by super merges.

\subsection{Combining Marks and Unicode Normalization}

The \texttt{T-2}, \texttt{F-2} and \texttt{P-1} word patterns all have a flaw in the handling of combining marks in the \lstinline|\p{M}| class.  These patterns only match letters, which then end up in a separate pretoken from any combining marks modify that letter.  For example, \'e, which can be represented as \lstinline{'e\u0301'} becomes two separate pretokens.  The \texttt{BOUNDLESS\_PATTERN} patterns keep all combining marks with their base character.  The pattern \lstinline|r"(?:\p{L}\p{M}*)"|, for example, is a letter and zero or more combining marks. These are wrapped in a non-capturing group, so the unit can be treated like a single character.  Combining marks are possible with all other Unicode classes, so this approach is used all across the \texttt{BOUNDLESS\_PATTERN} patterns. The final branch \texttt{B-11} of \lstinline|r"\p{M}+"| matches combining marks without a base letter. For example the combining acute accent \lstinline{'\u0301'} by itself is a valid Unicode code point, but is linguistically ill-formed without a base character.

The other approach to fix this problem is Unicode normalization. The \'e can also be written as a single pre-composed character \lstinline{'\u00E9'}.  Unicode normalization can be used to convert between these forms. \citet{gorman2025donttouchdiacritics}  show that normalization can fix this type of problems with combining marks. However, \citet[][App. D]{pmlr-v235-dagan24a} argue against normalization, as it is usually non-reversible.  Our approach of always keeping combining marks with their base characters solves the problem in a reversible way.  

\citet{{velayuthan-sarveswaran-2025-egalitarian}} present a more general problem with extended grapheme clusters in Tamil, Sinhala, and Hindi being broken up by pre-tokenization.  Some (but not all) graphemes are formed by a base character and one or more combining marks, so this would keep some of their graphemes intact. 

\subsection{Code Related Pretokens}

Patterns \texttt{B-2} to \texttt{B-6} collectively provide a new form of token alignment for variable and function names in code.  Programmers follow strict case conventions that allow the patterns to find the individual words within the names, in a programming language independent way.  The example \lstinline{'XMLHttpRequest snake_case camelCase CONSTANT'} becomes:

\lstinline{['XML', 'Http', 'Request', ' snake', '_case', ' camel', 'Case', ' CONSTANT'].}

Note that these are aligned with parts of the names.  With regular BPE, breaking variable names up in pre-tokenization would be prevent the full names from becoming tokens.  However, with superword merges these more aligned pretokens have the opportunity to recombine and form the complete variable names. Less common variable names will become several tokens. This is one example of how having supermerges allows more extensive pretoken alignment.

\subsection{Whitespace}
The existing whitespace regex (T-5 to T-6, F-5 to F-7, and P-4 to P-6) have small variations, but all use negative lookahead to cause the match to backtrack one space. 

With the example \lstinline{'Hello    world    \n\n  \n   '} and \texttt{GPT4\_PATTERN} we get:

\lstinline{['Hello', '   ', ' world', '    \n\n  \n', '   '].}

The \texttt{F-6} pattern matches the first three of the four spaces between \texttt{Hello} and \texttt{world}.  Then \texttt{F-2} picks off the third space in \texttt{' world'}.  The \texttt{F-5} matches multiple lines of whitespace ending in a line-ending, and then finally \texttt{F-6} picks up the remaining trailing whitespace. 

Breaking the last space off from longer runs of whitespace increases the number of words preceded by a space.  However, runs of multiple spaces are often encountered in the context of code.  Especially for space sensitive languages like python, splitting four spaces into three and one would seem to increase the difficulty of coding tasks.  Similarly, having white space span multiple lines, as in this example, seems to make coding tasks more difficult. So in contrast to the other approaches, \texttt{B-9} keeps runs of whitespace together, and breaks multiple lines of whitespace into separate tokens with one or more line endings.  For the example we would have:

\lstinline{['Hello', '    ', 'world', '    \n\n', '  \n', '   '].}

This is an area that would benefit from further experimentation. 

\subsection{Numbers}

\citet{singh2024tokenizationcountsimpacttokenization} take a detailed look at the tokenization of numbers.  Early models used patterns like \texttt{T-3}, that just used whichever numbers were found by BPE.  As they describe, models then switched to runs of 1 to 3 digits \texttt{(F-3, P-2)} or to using single digits.  By adding commas to numbers in the input context to enforce right-to-left tokenization they saw a dramatic decrease in arithmetic errors.  However, this can be done directly with regex \texttt{B-7} without the need for inserting commas, so that \lstinline{'1234567'} becomes the pretokens \lstinline{['1', '234', '567']}.

\subsection{Contractions}

The \texttt{T-1} and \texttt{F-1} patterns match English specific contraction endings like \texttt{\textquotesingle ve} or \texttt{\textquotesingle ll}.  Disliking the English-specific nature, this was omitted from Punct.  Pattern \texttt{B-1} is more language-independent.  It keeps the ending of the contraction together with the word as a single pretoken. Thus, any word containing a straight or curly apostrophe internal to the word is matched, like \texttt{C\textquotesingle est} or \texttt{J\textquotesingle ai}. Patterns \texttt{O-1} and \texttt{O-2} keep the contraction ending with the word.

\section{Token distribution at additional vocabulary sizes}
\label{app:extra_vocab_sizes}

\begin{figure}[ht]
    \centering
    \includegraphics[width=\linewidth]{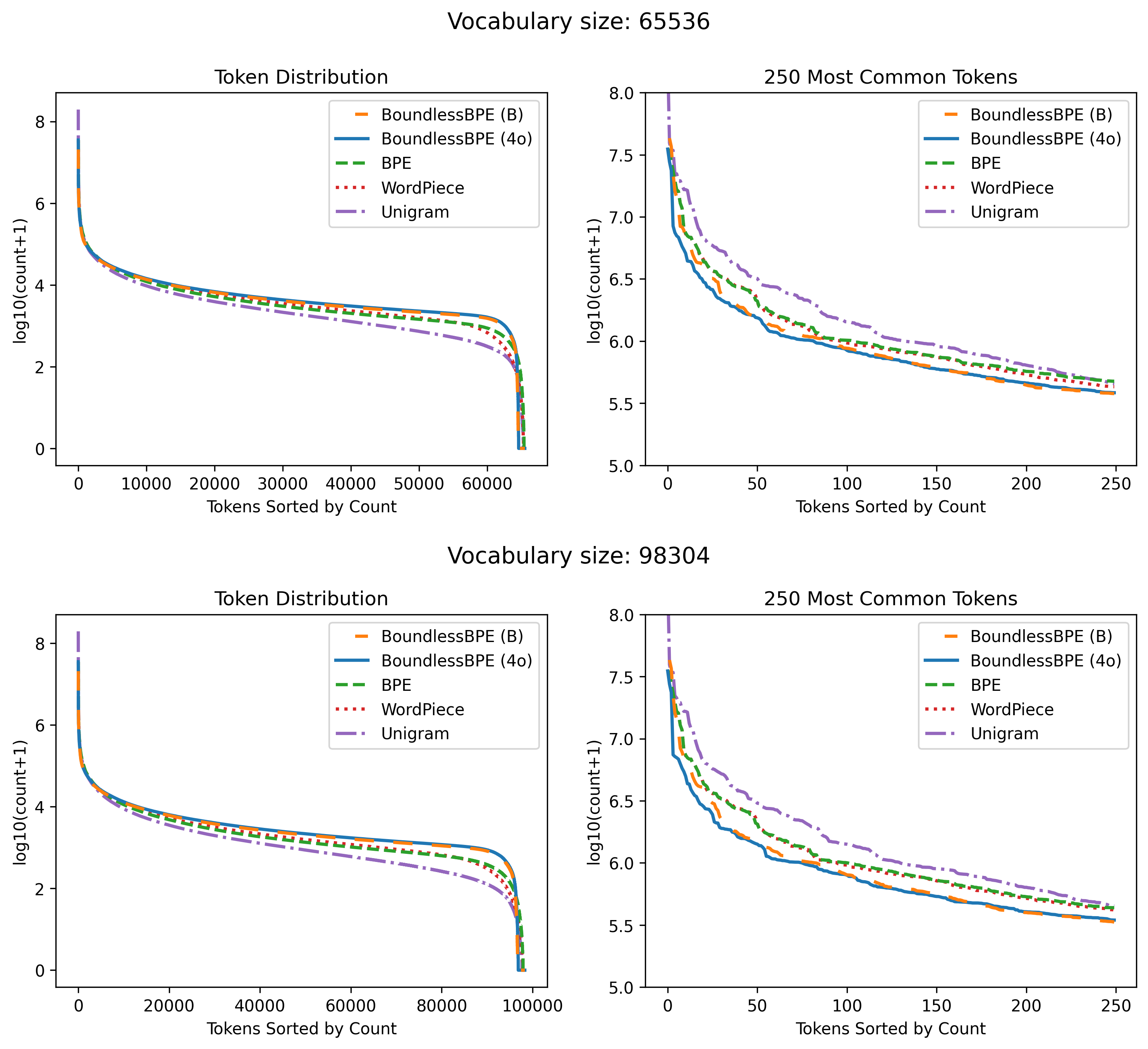}
    \caption{Left column is the $\log_{10}(\text{count}+1)$ for each token, sorted from most to least frequent on the $x$-axis on our evaluation corpus. The +1 is to allow plotting of 0 counts. Right column shows a zoomed-in view of the 250 most common tokens.}
    \label{fig:extra_vocab_sizes}
\end{figure}

\cref{fig:extra_vocab_sizes} shows the same plot as \cref{fig:token_distribution} at the vocabulary sizes of 65,536 and 98,304.  The trends are largely the same as in \cref{fig:token_distribution}, with \ourmethod{} having higher counts at the low frequency end of the distribution compared to baselines, and lower frequencies for the most common tokens.  Both these are desirable to have a more uniform distribution.

\section{Efficient implementation}
\label{app:efficient_implementation}
In the representation shown in \cref{sub:inference_example}, a document is divided into pretokens, each of which contains one or more tokens. This is simple conceptually and can be used directly for training, but is extremely slow. Tokenizer training routines commonly use the trick of aggregating the pretokens produced over a large training corpus, and use the aggregate counts when calculating pairwise merge counts. Thus the pretoken \lstinline{[' the']} is only ever tokenized once even though it might appear hundreds of thousands of times. This speedup is crucial for performance reasons. \footnote{We have our own implementation of BPE with deletions using this speedup, based on Andrej Karpathy's \texttt{minbpe} library. \url{https://github.com/karpathy/minbpe}}

This allows a faster implementation of \ourmethod{}.  We keep two separate sets of the training data.  We pre-tokenize and tally up the frequency of the pretokens for regular merges.  The aggregation here results in more than a 10x speedup on the training time. This first set of data is initialized with single bytes, and merges are selected according to the total aggregated counts.  

We have a separate second copy of the training data for the supermerges, where each document is initially broken into pretokens, which will combined into superwords. Since most documents are distinct, no aggregation can help speed up supermerges. 

These two representations are related through the process of \emph{unlocking} pretokens.  The first subword representation proceeds as in regular BPE.  However, when a pretoken there is reduced to a single token, we say that that pretoken has been unlocked.  

The second superword representation can only consider a pairwise merge when two adjacent words are unlocked.  After a new pretoken is unlocked, the counts of the superword must be updated accordingly. We track the pairwise and single counts of each representation.  For speed, we dynamically compute the changes in counts that result from performing each merge. Despite these optimizations, it still took 4.7 CPU days to train our 1GB training dataset due to the lack of aggregation for supermerges. For comparison, training Hugginface BPE on the same data, vocabulary size, and machine takes 59s.

\section{Metrics for token distribution uniformity}
\label{app:renyi_analysis}

\begin{figure}[ht!]
    \centering
    \includegraphics[width=\linewidth]{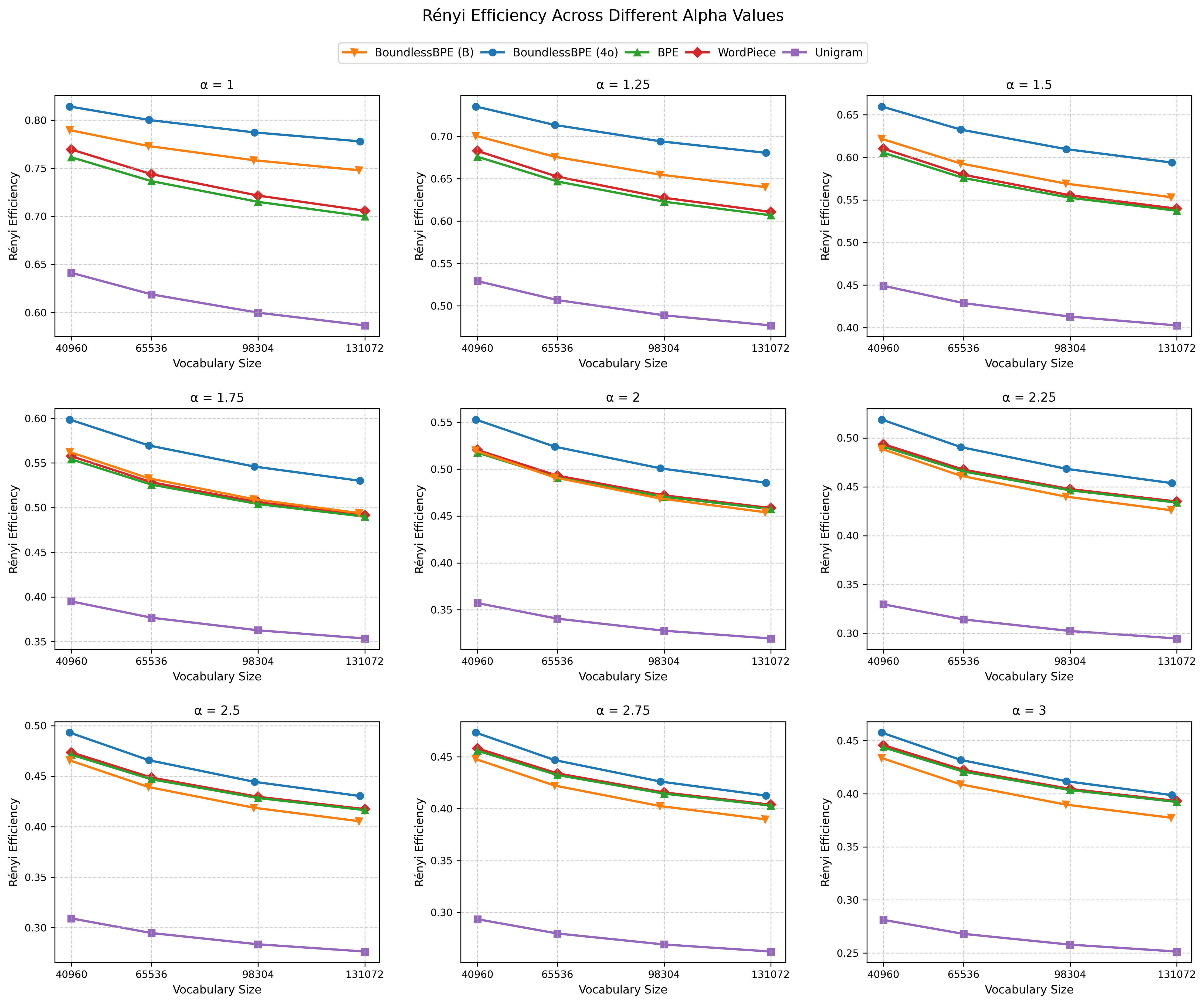}
    \caption{R\'enyi efficiency at various values of $\alpha$}
    \label{fig:renyi_entropy_grid}
\end{figure}

\citet{zouhar-etal-2023-tokenization} suggest the use of R\'enyi efficiency as a metric to measure token distribution uniformity. The equation contains a hyperparameter $\alpha \ge 1$ which controls the exponent on each proportion.  They suggest the use of $\alpha=2.5$, which is in the range that strongly penalizes very common tokens. \cref{fig:renyi_entropy_grid} show the relative R\'enyi efficiency for various values of $\alpha$.  The improvement of \ourmethod{} (4o) over the baselines increases at lower values of $\alpha$. \ourmethod{} (B) outperforms the baselines at lower values of $\alpha$.

\section{Ablation results at additional vocabulary sizes}
\label{app:ablation_extra_vocab_sizes}

\cref{tab:more_ablation} shows the ablation results at the vocabulary sizes of 65,536 and 98,304.

\begin{table}[ht!]
\centering
\begin{tabular}{cllcc}
\toprule
\textbf{Vocab Size} & \textbf{Regex Pattern} & \textbf{PickyBPE} & \textbf{Bytes Per Token} & \textbf{R\'enyi Efficiency} \\
\midrule
65,536 & Boundless & none & 4.488 & 0.4386 \\
65,536 & Boundless & original & 4.486 & 0.4386 \\
65,536 & Boundless & ours & 4.480 & 0.4387 \\
\midrule
65,536 & GPT4o & none & 4.673 & 0.4652 \\
65,536 & GPT4o & original & 4.671 & 0.4653 \\
65,536 & GPT4o & ours & 4.664 & 0.4653 \\
\midrule
98,304 & Boundless & none & 4.660 & 0.4181 \\
98,304 & Boundless & original & 4.658 & 0.4182 \\
98,304 & Boundless & ours & 4.651 & 0.4183 \\
\midrule
98,304 & GPT4o & none & 4.871 & 0.4439 \\
98,304 & GPT4o & original & 4.869 & 0.4439 \\
98,304 & GPT4o & ours & 4.861 & 0.444 \\
\bottomrule
\end{tabular}
\caption{Effect of regular expression pattern and PickyBPE style on intrinsic measures for vocabulary sizes of 65,536 and 98,304}
\label{tab:more_ablation}
\end{table}

\section{Additional related work}
\label{app:additional_related_work}
\paragraph{\textbf{Foundational Subword Tokenization Algorithms}} Byte level subword tokenization has become a fundamental component in modern NLP, balancing vocabulary size with morphology and out-of-vocabulary handling. Byte Pair Encoding (BPE; \citeauthor{sennrich-etal-2016-neural}, \citeyear{sennrich-etal-2016-neural}) iteratively merges the pair of adjacent tokens with the highest count to build a vocabulary. WordPiece \citep{wordpiece-org} is similar to BPE, except merges are selected according to their Pointwise Mutual Information (PMI). \citet{kudo-2018-subword} introduced the UnigramLM tokenizer, a top-down approach that starts with a large vocabulary and prunes tokens based on their contribution to sequence likelihood according to a unigram language model.

\paragraph{\textbf{Removal of Intermediate Tokens}}
The pairwise nature of BPE merges mean some \textit{scaffold} tokens added to the vocabulary simply serve as a bridge to a more popular token, but are not often used on their own. \citet{bostrom-durrett-2020-byte} observed a dead zone of such tokens in both UnigramLM and BPE vocabularies. The proportion of such tokens were found to be higher in the case of BPE, motivating research into their removal and vocabulary refinement. In \citet{lian2024scaffoldbpeenhancingbytepair} scaffold tokens are marked during a training process, and split into components as a postprocessing step during inference. \citet{chizhov-etal-2024-bpe} integrate the deletion step into the BPE training process, where it can affect later merge decisions. Thus an ordered combination of merges and deletions occur during inference. We find \citet{chizhov-etal-2024-bpe}'s approach particularly compelling due to its direct integration into BPE training. In contrast, \citet{bauwens-delobelle-2024-bpe} focused on a post-processing step to remove merge rules decreasing morphological alignment. We opted against a purely post-processing approach like \citet{cognetta-etal-2024-analysis} to maintain tighter control over vocabulary construction during training.

\end{document}